%% file: main.tex
\documentclass{svproc}
\usepackage{url}
\usepackage[numbers, sort&compress]{natbib}

\usepackage[utf8]{inputenc} 
\usepackage[T1]{fontenc}    
\usepackage{hyperref}       
\usepackage{url}            
\usepackage{booktabs}       
\usepackage{amsfonts}       
\usepackage{nicefrac}       
\usepackage{microtype}      
\usepackage{graphicx}
\usepackage{float}
\usepackage{amsmath}
\usepackage{subcaption}
\usepackage{array}
\usepackage{dblfloatfix}
\usepackage{mwe}
\usepackage{graphbox}

\begin{document}
\mainmatter              
\title{CanvasGAN: A simple baseline for text to image generation by incrementally patching a canvas}
\titlerunning{CanvasGAN}  
%
\author{Amanpreet Singh and Sharan Agrawal}
\authorrunning{Singh et al.} 
%
\tocauthor{}
\institute{New York University}

\maketitle              

\begin{abstract}
 We propose a new recurrent generative model for generating images from text captions while attending on specific parts of text captions. Our model creates images by incrementally adding patches on a "canvas" while attending on words from text caption at each timestep. Finally, the canvas is passed through an upscaling network to generate images. We also introduce a new method for generating visual-semantic sentence embeddings based on self-attention over text. We compare our model's generated images with those generated  \citet{reed2016generative}'s model and show that our model is a stronger baseline for text to image generation tasks.
\end{abstract}

\input{sections/intro.tex}
\input{sections/related.tex}

\input{sections/model.tex}

\input{sections/experiments.tex}
\input{sections/conclusion.tex}

\bibliographystyle{plainnat}
\bibliography{bibliography}
\newpage
\input{appendices/attention_weights.tex}
\end{document}

%% file: sections/intro.tex
\section{Introduction}
\label{sec:introduction}
With introduction of Generative Adversarial Networks (GAN) \citet{goodfellow2014generative} and recent improvements in their architecture and performance [\citenum{arjovsky2017wasserstein}][\citenum{gulrajani2017improved}], the focus of research community has shifted towards generative models. Image generation is one of the central topic among generative models. As a task, image generation is important as it exemplifies model's understanding of visual world semantics. We as humans take context from books, audio recordings or other sources and are able to imagine corresponding visual representation. Our models should also have same semantic understanding of context and should be able to generate meaningful visual representations of it. Recent advances, quality improvements and successes of the discriminative networks has enabled the industrial applications of image generation [\citenum{NIPS2016_6527}] [\citenum{isola2016image}] [\citenum{CycleGAN2017}]. In this paper, we propose a sequential generative model called CanvasGAN for generating images based on a textual description of a scenario. Our model patches a canvas incrementally with layers of colors while attending over different text segments at each patch.

Variety of sequential generative models have been introduced recently, which were shown to work much better in terms of visual quality as model gets multiple chances to improve over previous drawings. Similar to CanvasGAN's motivation from human execution of painting, attention is motivated by the fact that we humans improve our performance by focus on a particular aspect of task at a moment rather than whole of it [\citenum{bahdanau2014neural}][\citenum{xu2015show}]. In recent works, attention alone has been shown to be really effective without anything else \citep{vaswani2017attention}. For our model, at each timestep, model focuses on a particular part of text for creating a new patch rather than whole sentence.

A lot of models have been proposed in the recent year for the text to image generation task [\citenum{zhang2016stackgan}][\citenum{reed2016generative}][\citenum{attngan2017}]. Many of these models incorporate GANs and a visual semantic embedding \citep{frome2013devise} for capturing text features which are semantically important to images. Generator networks create an image from sampled noise conditioned on the text and discriminator predicts whether the images is real or generated. However, the images generated by these networks are mostly not coherent and seem distant from text's semantics. To overcome this incoherence, we propose new method for generating image-coherent sentence embeddings based on self-attention over the text.

In this work, we make two major contributions:
\begin{enumerate} \itemsep -3pt
    \item We propose a new model for text to image generation, called CanvasGAN, which analogous to human painters generates an image from text incrementally by sequentially patching an empty canvas with colors. Furthermore, CanvasGAN uses attention to focus over text to use for generating new patch for canvas at each time-step.
    \\
    \item We introduce a new visual semantic embedding generation mechanism which uses self-attention to focus on important hidden states of RNN to generate a sentence embedding instead of taking hidden state at last timestep as usual. This sentence embedding generator is separately trained to be coherent with image semantics using pairwise ranking loss function between sentence embedding and image. 
\end{enumerate}

%% file: sections/related.tex
\section{Related work}
\label{sec:related_work}
Generative Adversarial Networks (GAN) \cite{goodfellow2014generative} and Variational Auto-Encoders (VAE) \cite{kingma2013auto} can be considered as two major categories in deep generative models. In conditional image generation, GANs have been studied in-depth where the initial work used simple conditional variables like object attributes or class labels (MNIST) to generate images [\citenum{van2016conditional}][\citenum{odena2016conditional}][\citenum{yan2016attribute2image}]. Multiple models were introduced in image to image translation which encompasses mapping from one domain to other [\citenum{CycleGAN2017}][\citenum{kim2017learning}][\citenum{zhu2017toward}], style transfer [\citenum{gatys2016image}] and photo editing [\citenum{brock2016neural}][\citenum{zhu2016generative}].



In context of sequential generative models, \citet{denton2015deep} uses a laplacian pyramid generator and discriminator called LAPGAN to synthesize images from low to high resolutions levels sequentially. Similar to our work, DRAW network \citenum{gregor2015draw} is a sequential version of an auto-encoder where images are generated by incrementally adding patches on a canvas. Closest to our work is \citet{mansimov2015generating}, which uses a variational auto-encoder to patch a canvas and then use an inference network to map back to latent space. In CanvasGAN, we use a discriminator based loss function with GAN based architecture and our new visual-semantic embedding to generate images from text.

\begin{figure}[t]
  \includegraphics[width=\textwidth]{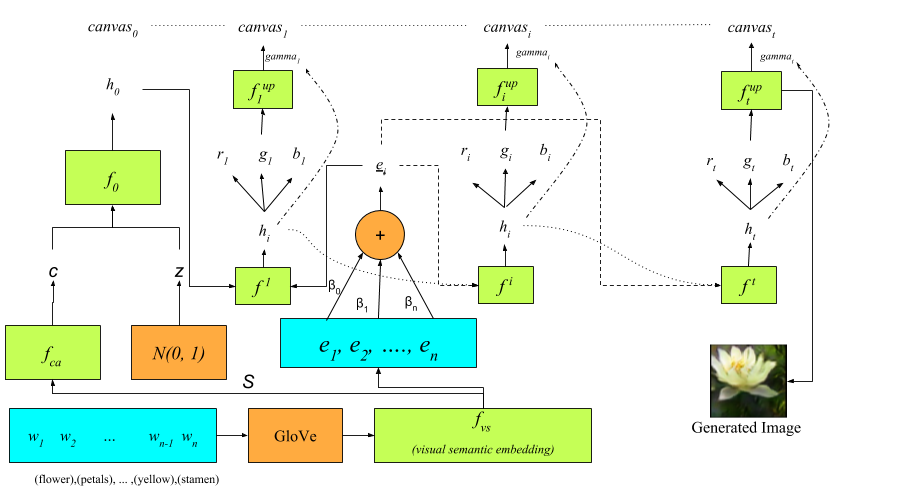}
  \caption{CanvasGAN for generating images from text through incremental patches on an empty canvas. Starting from an empty canvas, at each timestep, we attend over text features to get relevant features. These features are passed as input to a recurrent unit which uses hidden state from previous timestamp to figure out what to patch. Finally, through upsampling the hidden state, we get the three channels which are finally combined and patched to the canvas based on $\gamma_i$.}
\label{fig:generator}
\end{figure}

Image caption generation, which is reverse of text to image generation has seen a lot of significant models which have good performance. \citet{karpathy2015deep} uses alignments learned between CNN over image regions and BiRNN over sentences through multi-modal embedding to infer descriptions of image regions. \citet{xu2015show} uses attention over weights of a convolutional net's last layer to focus on a particular area of image to generate captions as a time-series data via a gated recurrent unit [\citenum{hochreiter1997long}][\citenum{chung2014empirical}]. 

In \citet{reed2016generative}, a simple generative model was introduced for converting text captions into images. This model started a series of work on text-to-image generation task. It contained a simple upscaling generator conditioned on text followed by a downsampling discriminator. The model also used visual semantic embedding \cite{reed2016learning} for efficiently representing text in higher continuous space which is further upsampled into an image. This was further improved by \citet{zhang2016stackgan}, by generating high quality 256x256 images from a given text caption. Their model (StackGAN) employs a two step process. They first use a GAN to generate a 64x64 image. Then, this image along with text embedding is passed on to another GAN which finally generates a 256x256 image. Discriminators and generators at both stages are trained separately. 

Most notable recent works in the text to image task are AttnGAN \cite{attngan2017} and HDGAN \cite{Zhang2018PhotographicTS}. In AttnGAN, the authors improved StackGAN by using attention to focus on relevant words in the natural language description and proposed a deep attentional multimodal similarity model to compute a fine-grained image-text matching loss for training the visual semantic embedding generator. In HDGAN, authors generate images at different resolutions with a single-streamed generator. At each resolution, there is a separate discriminator which tells (i) whether image is fake or real and (ii) if it matches the text or not. Another important contribution is made by \citet{selfattngan2018} in which the authors generate details using cues from all feature locations in the feature maps. In their model (SAGAN), the discriminator can check that highly detailed features in distant portions of the image are consistent with each other which leads to a boost in the inception score.

%% file: sections/model.tex
\section{Model}

\label{sec:model}
We propose CanvasGAN, a new sequential generative network for generating images given a textual description. Model structure is shown in Figure \ref{fig:generator}. We take motivation from human painters in how they create a painting iteratively instead of painting it in single step. Keeping that in mind, starting with an empty canvas, we paint it with patches iteratively. At each step, a patch is generated based on attended features from text.

First, we retrieve GloVe \cite{pennington2014glove} embeddings $g$ for the words in caption, $w$ and encode it using our visual semantic network $f^{vs}$ which we explain in Section \ref{subsec:visual_semantic_embedding}. 
This provides us with sequentially encoded word embeddings, $e$ and a sentence embedding, $s$ for whole sentence. We sample our noise $z \in \mathbb{R}^D$ from standard normal distribution $\mathcal{N}\left(0, 1\right)$. 
We use conditional augmentation, $f^{ca}$ over sentence embedding, $s$ to overcome the problem of discontinuity in latent mapping in higher dimensions due to low amount of data [\citenum{zhang2016stackgan}]. 
Conditional augmented sentence vector $c$ along with noise $z$ is passed through neural network $f$ to generate initial hidden state, $h_0$ of the recurrent unit. 
Initially, canvas is empty and a zero tensor $canvas_0 = \vec{0}$. Now at each timestep, $i \in \left\{1, \ldots, t\right\}$, we execute a series of steps to patch canvas. 
Attention weights $\alpha_i$ are calculated by neural network $f_i^{att}$ with inputs $c, z$ and $h_{i - 1}$ which are scaled between (0, 1) as $\beta_i$ by taking a softmax. 
Attended sentence embedding, $\bar{e_i}$ for current time-step is calculated as $\sigma_j\beta_{ij}e_j$. 
Next hidden state, $h_i$ is calculated by recurrent unit $f^i$ which take previous hidden state $h_{i - 1}$ and attended sentence embedding $\bar{e_i}$ as inputs. 
The $r, g$ and $b$ channels for next patch are produced by neural networks $f_i^r, f_i^g$ and $f_i^b$ which are concatenated to produce a flattened image. 
This image is passed through an upscaling network, $f_i^{up}$, to generate a patch of size same as canvas. 
For adding this patch to canvas, we calculate a parameter $\gamma$ from neural network, $f_i^{\gamma}$ which determines how much of $delta$ will be added to the canvas. Finally, $\gamma * \delta$ is added $canvas_{i - 1}$ to generate $canvas_i$. Our final image representation is denoted by $canvas_t$ at last time-step. Mathematically, our model can be written as,
\begin{gather*}
canvas_0 = \vec{0} \\
z \sim \mathcal{N}(0, 1), g = GloVe(w) \\
e, s = f^{vs}(g) \\
c = f^{ca}(s) \\
h_0 = f_0(c, z) \\
\end{gather*}
for $i = 1, \ldots, t$:

\begin{gather*}
\alpha_i = f^{att}_i(c, z, h_{i - 1}) \\
\beta_i = \frac{\text{exp}(\alpha_i)}{\sum_j\text{exp}(\alpha_{ij})}, \bar{e_i} = \sum_j\beta_{ij}e_j\\
h_i = f^{i}(\bar{e_i}, h_{i - 1}) \\
r_i = f_i^r(h_i), g_i = f_i^g(h_i), b_i = f_i^b(h_i), \gamma = f_i^{\gamma}(h_i) \\
\delta = f_i^{up}(\left[r;g;b\right]) \\
canvas_i = canvas_{i-1} + \gamma * \delta
\end{gather*}

For our experiments, we implement $f_{ca}$, $f_i^r$, $f_i^g, f_i^b$ and $f_i^\gamma$ as simple affine network followed by rectified linear unit non-linearity. We choose $f_0$ and $f_i$ as Gated Recurrent Unit [\citenum{chung2014empirical}]. For $f^{att}_{i}$, we first concatenate $c, z$ and $h_{i-1}$ and then, pass them through a affine layer followed by a softmax. Finally, upscaling network using deconvolutional layers with residual blocks to scale the image to higher resolutions  [\citenum{lecun1995convolutional}][ \citenum{zeiler2010deconvolutional}].


\begin{figure*}
    \includegraphics[width=\textwidth]{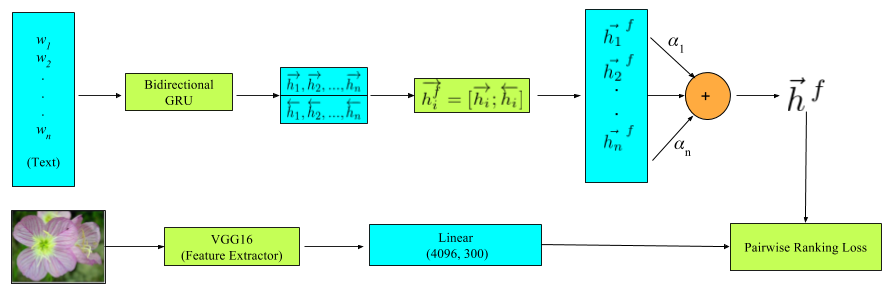}
  \caption{Architecture of self-attended visual semantic embedding. Word representations ($\protect\overrightarrow{w_i}$) from text-embeddings and passed through a bidirectional recurrent unit whose forward ($\protect\overrightarrow{h_i}$) and backward ($\protect\overleftarrow{h_i}$) 
  hidden states are summed to get one hidden state ($\protect\overrightarrow{h^f_i}$) for each word. We pass these hidden states through a feed-forward network to calculate attention weights ($\protect\alpha_i$) for each hidden state. Finally, we linearly combine hidden state multiplied by their respective attention weights to get final hidden state. This is compared via pairwise ranking loss with image representation of original image downsampled through a CNN to same number of dimensions. Whole network is trained end-to-end.}
  \label{fig:visual_semantic_embedding}
\end{figure*}

\subsection{Visual Semantic Embedding}
\label{subsec:visual_semantic_embedding}
Standard models for representations of words in continuous vector space such as $GLoVe$ [\citenum{pennington2014glove}], $Word2Vec$ [\citenum{mikolov2013efficient}] or $fastText$ [\citenum{joulin2016bag}] are trained on a very large corpus (e.g. Wikipedia) which are not visually focused. 
To overcome this issue, visual semantic models have been proposed in recent works [\citenum{reed2016learning}][\citenum{kiros2014unifying}]. These representations perform well, but don't have power to focus on important words in text. In these embeddings, focus is on last encoded state of recurrent unit as in machine translation[\citenum{bahdanau2014neural}]. We propose a new method for calculating visually semantic embedding using self-attention. By using self-attention over encoded hidden states of a recurrent units, our method generates a sentence embedding in latent space. To introduce component of visual semantic, we compare image with sentence embedding using pairwise ranking loss function similar to \citep{kiros2014unifying}. For proper comparison, image is encoded into same latent space as text by passing features extracted from last average-pooling layer of pretrained Inception-v3 through an affine layer. This method allows embedding to focus on most important part of sentence to generate sentence embedding.
To introduce visual semantic component, a lot of models were proposed. Some of these models used a smaller datasets with visual descriptions, like Wikipedia article's text along with its images. These embeddings didn't perform well in practice. Also, these models were not capable of being generalized for any zero-shot learning task and had to be fine-tuned separately according to the task, To overcome this, a lot of visual semantic models [\citenum{reed2016learning}][\citenum{kiros2014unifying}] were proposed which take in account both the image and its representative text while training.  

\begin{figure*}
  \includegraphics[width=\textwidth]{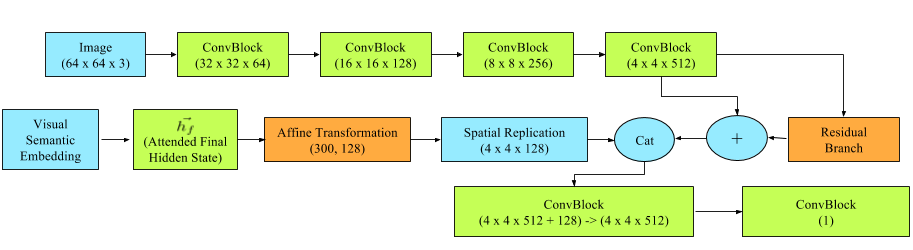}
  \caption{Architecture of our model's discriminator. Image is downsampled and passed through residual blocks to extract feature. Then, it is concatenated with spatially replicated text representation and the combination is passed through convolutional layers to predict the probability of whether the text is semantically relevant to image or not.}
\label{fig:discriminator}
\end{figure*}

\begin{gather}
\overrightarrow{h_1}, \overleftarrow{h_1}, \overrightarrow{h_2} , \overleftarrow{h_2}, \ldots, \overrightarrow{h_n} , \overleftarrow{h_n} = GRU(\overrightarrow{w_1}, \overrightarrow{w_2}, \ldots, \overrightarrow{w_n}) \\
\overrightarrow{h^f_i} = \overleftarrow{h_i} + \overrightarrow{h_i}  \\
\alpha_i = f_{score}(h^f_i) \\
h_f = \sum_i^n\alpha_ih^f_i
\end{gather}

where $\overrightarrow{w_1}$, $\overrightarrow{w_2}$, $\ldots$, $\overrightarrow{w_n}$ are the vector representation of \textbf{y}, original words in one-hot (1-of-K) encoded format, \textbf{y} = ($y_1$, $y_2$, \ldots, $y_n$) where $K$ represents the size of vocabulary and $n$ is length of sequence. GRU is our gated recurrent unit, while $f_{score}$ is scoring function for attention weights ($\alpha_i$). $\overrightarrow{h_i}$ and $\overleftarrow{h_i}$ are bidirectional hidden states for $\overrightarrow{w_i}$. $\overrightarrow{h^f_i}$ is sum of both bidirectional hidden states which is combined with $\alpha_i$ to get final hidden state $h_f$

\subsection{Discriminator}
\label{subsec:discriminator}
Our model's discriminator is a general downsampling CNN which takes an image and a text representation to predict whether image is semantically relevant to text or not. Image is downsampled and passed through residual branch for feature extraction. Text representation ($h_f$) is spatially replicated so that it and the image are of same dimensionality. These are concatenated and further passed through convolutional layers to generate a real number between 0 and 1 predicting the probability of semantic relevance.

%% file: sections/experiments.tex
\section{Experiments and Results}
\label{sec:experiments}
We train our model on oxford flowers-102 dataset \cite{nilsback2008automated} which contains 103 classes for different flowers. Each flower has an associated caption each describing different 
aspects, namely the local shape/texture, the shape of the boundary, the overall spatial distribution of petals, and the colour. 
\begin{figure*}[h]
  \centering
  \includegraphics[width=\textwidth]{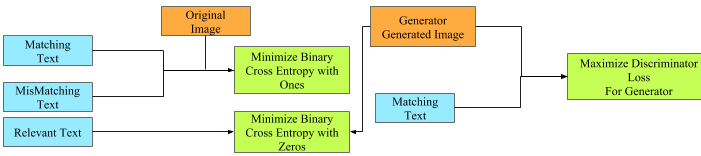}
  \caption{Calculation of Generator and Discriminator loss using matching, mismatching and relevant text. In the left side of the figure, we optimize our discriminator using matching text and mismatching text with the original training image. We also use relevant text with generator generated images to further optimize the discriminator. On the right side of the figure, we use discriminator's loss for generator generated image for matching text and try to maximize that loss.}
\label{fig:losses}
\end{figure*}

We create extra augmented data via rolling matching text across the text dimension which matches each text with wrong image and thus creates a mismatching batch corresponding to a matching batch. We also create a batch of relevant text in which we roll half of the text with rest of the batch, in this case the text is somewhat relevant to image but still doesn't match semantically. We directly pass mismatching and matching text and images to discriminator while minimizing binary cross entropy (BCE) loss with one (relevant) and zero (non-relevant) respectively. For relevant text, we pass the text to generator and further pass generated image to discriminator while minimizing BCE loss with zero (fake). We also calculate generator loss in case of matching text by passing generated image to discriminator with task and minimizing BCE loss for discriminator prediction with respect to zero. See Figure \ref{fig:losses} for overview of loss calculation.

\subsection{Quantitative Analysis}
\begin{figure*}[t]
\begin{subfigure}{0.33\textwidth}
  \centering
  \includegraphics[width=1\linewidth]{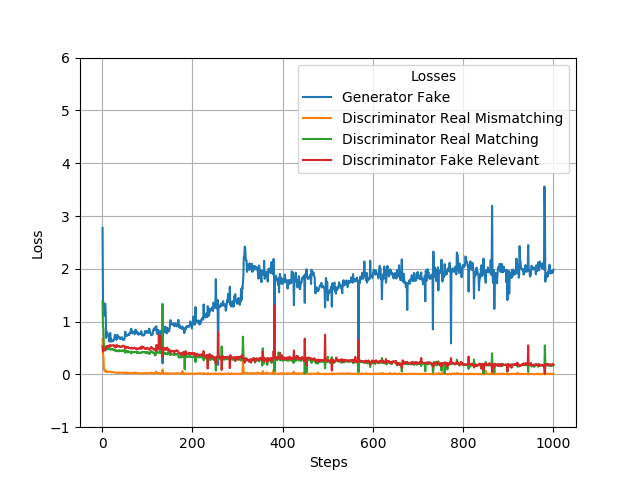}
  \caption{}
  \label{fig:sfig1}
\end{subfigure}%
\begin{subfigure}{0.33\textwidth}
  \centering
  \includegraphics[width=1\linewidth]{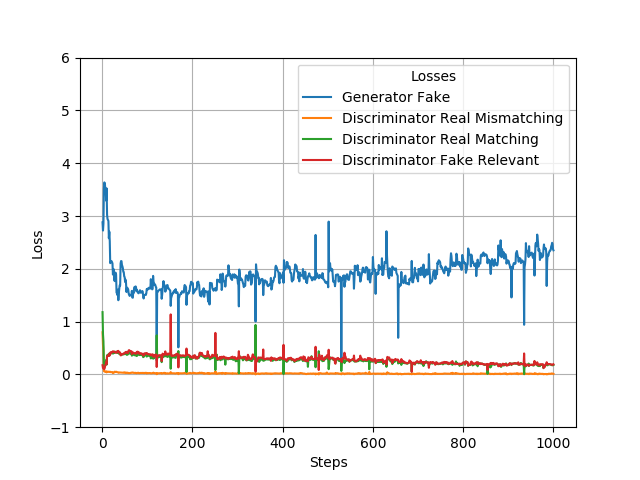}
  \caption{}
  \label{fig:sfig2}
\end{subfigure}
\begin{subfigure}{0.33\textwidth}
  \centering
  \includegraphics[width=1\linewidth]{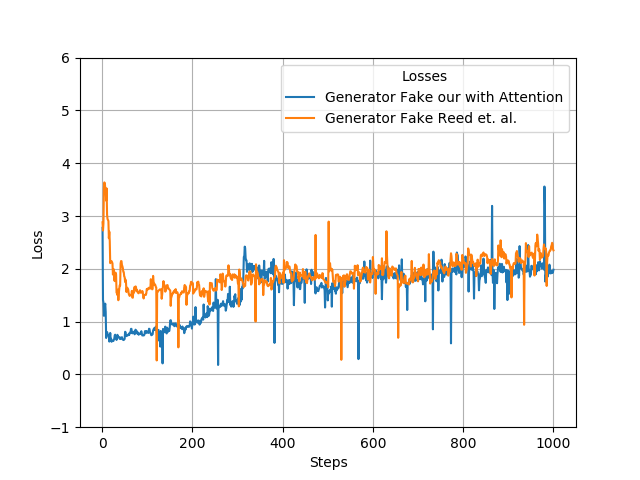}
  \caption{}
  \label{fig:sfig3}
\end{subfigure}
\caption{Loss vs steps curves are shown for ours and \citet{reed2016generative} model. Figure \ref{fig:sfig1} shows loss curves for RCAGAN which includes Generator loss, Discriminator loss with matching, mismatching and relevant text. Figure \ref{fig:sfig2} shows similar loss curves for \citet{reed2016generative}. Figure \ref{fig:sfig3} shows comparison of Generator loss for RCAGAN and \citet{reed2016generative}}.
\label{fig:loss_curves}
\end{figure*}

For analyzing CanvasGAN quantitatively, we will review the loss curves for generator and different losses of discriminator. In Figure \ref{fig:loss_curves}, we can see various loss curves for generator and discriminator losses for both ours and \citet{reed2016generative} model. In both models, discriminator loss for matching and relevant text drops progressively with time and shows that discriminator gets better. For relevant text though, discriminator loss drops close to zero in the beginning as the generator is untrained and is not able to generate plausible images. However, it recovers from that after a few epochs. 

For CanvasGAN's loss curves, we can see the evident effect of applying RNN based incremental generator with attention: generator loss drops quite drastically in the start as discriminator fails to catch up with the generator. This shows that with attention, the generator is able to produce plausible images from the get-go. After a few epochs, the discriminator eventually copes up with the generator and the generator loss starts increasing, which should result in a better discriminator and generator overall and it is supported by the fact that our loss always remains below than that of \citet{reed2016generative}.  

\begin{figure*}[t]
    \includegraphics[width=\textwidth]{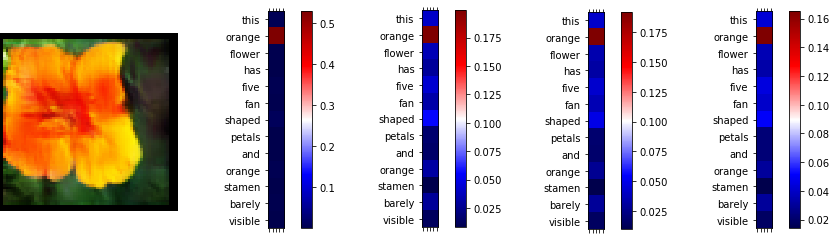}
    \caption{Attention weights for 4 timesteps for image caption "this orange flower has five fan shaped petals and orange stamen barely visible". We can see how the attention weights change with timesteps. Significant weights are concentrated around "orange". The image on left is the generated image.}
    \label{fig:attention_weights}
\end{figure*}
\begin{table}[h!]
    \begin{tabular}{p{4cm}ccc}
    \hline
    Caption & Reed et al & CanvasGAN & Original\\
    \hline
    this flower has white petals with pointed tips and a grouping of thick yellow stamen at its center. &
    \includegraphics[align=t,width=0.3\textwidth]{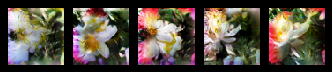} &\includegraphics[align=t,width=0.3\textwidth]{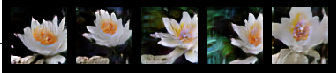} & \includegraphics[align=t,width=0.061\textwidth]{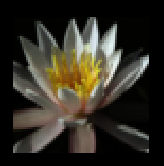}\\
    \hline 
    this flower has large pink petals with a deep pink pistil with a cluster of yellow stamen and pink pollen tubes &\includegraphics[align=t,width=0.3\textwidth]{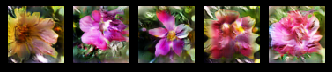} &\includegraphics[align=t,width=0.3\textwidth]{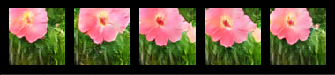} 
    &\includegraphics[align=t,width=0.061\textwidth]{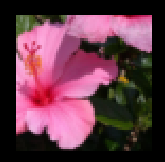}\\
    \hline 
    this flower has petals that are red and has yellow stamen &\includegraphics[align=t,width=0.3\textwidth]{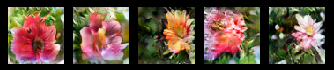} &\includegraphics[align=t,width=0.3\textwidth]{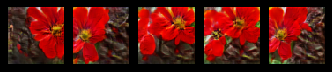} 
    &\includegraphics[align=t,width=0.061\textwidth]{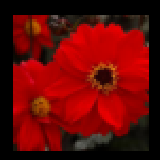}\\
    \hline 
    the violet flower has petals that are soft, smooth and arranged separately in many layers around clustered stamens &\includegraphics[align=t,width=0.3\textwidth]{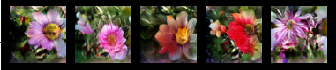} &\includegraphics[align=t,width=0.3\textwidth]{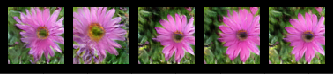} 
    &\includegraphics[align=t,width=0.061\textwidth]{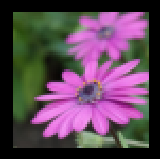}\\
    \hline
    there is a star configured array of long thin slightly twisted yellow petals around an extremely large orange and grey stamen covered ovule &\includegraphics[align=t,width=0.3\textwidth]{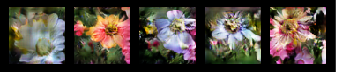} &\includegraphics[align=t,width=0.3\textwidth]{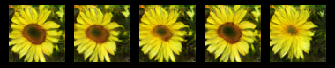} 
    &\includegraphics[align=t,width=0.061\textwidth]{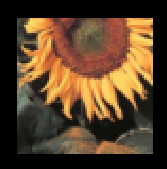}\\ 
    \hline
    \end{tabular}
    \caption{Comparison of images generated by our CanvasGAN model and those generated by \citet{reed2016generative}. We also provide original image related to caption for reference.}
    \label{table:gen_images}
\end{table}
Figure \ref{fig:attention_weights} shows the attention weights calculated per timestep for an image caption. Our results show that RCA-GAN almost always focuses on color and shape mentioned in image caption which is very good for visual semantics. For more attention weights maps, see Appendix 1.

To evaluate CanvasGAN's generative capabilities, we calculated the inception score \cite{salimans2016improved} using pre-trained ImageNet. We achieved an inception score \cite{salimans2016improved} of 2.94 $\pm$ 0.18 using only 4 timesteps which is close to the score of 3.25, achieved by the state-of-the-art model \cite{zhang2016stackgan}. This shows that RCA-GAN has a huge potential--with further optimizations and increased timesteps, it can perform much better.

\subsection{Qualitative Analysis}
Our results show that images generated by CanvasGAN are always relevant to text and never non-sensical as is the case we observed with \citet{reed2016generative}. Table \ref{table:gen_images} shows the images generated by both models for a certain text description. 

We can see that CanvasGAN generates semantically relevant images almost all of the time, while \citet{reed2016generative} generates distorted relevant images most of the times, but fails badly on images with long captions. CanvasGAN's outputs that have been generated incrementally and then sharpened using CNN are usually better and expressive. Further improvements for quality can be made by generating more patches by increasing number of time-steps.

%% file: sections/conclusion.tex
\section{Conclusions}
Text to image generation has become an important step towards models which better understand language and its corresponding visual semantics. Through this task we aim to create a model which can distinctly understand colors and objects in a visual sense and is able to produce coherent images to show it. There has been a lot of progress in this task and many innovative models have been proposed but the task is far from being solved yet. With each step we move towards human understanding of text as visual semantics. In this paper, we propose a novel architecture for generating images from text incrementally like humans by focusing at a part of time at a particular incremental step. We use GAN based architecture using a RNN-CNN generator which incorporates attention and we name it CanvasGAN. We show how the model focuses on important words in text at each timestep and uses them to determine what patches to add to canvas. Finally, we compare our model with previous prominent work and show our generator's comparatively better results. 
\label{sec:conclusions}

%% file: appendices/attention_weights.tex
\section*{Appendix 1: Attention weights}
\label{appendix:1}
\begin{figure}[h]
\centering
\begin{subfigure}{1\linewidth}
  \centering
  \includegraphics[width=1\linewidth]{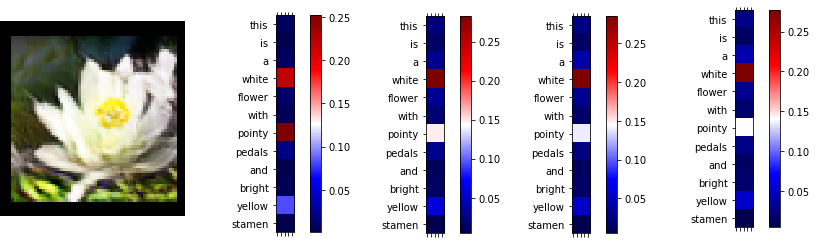}
  \caption{Text caption is: "this is a white flower with pointy pedals and bright yellow stamen"}
  \label{fig:sfig1}
\end{subfigure}

\begin{subfigure}{1\linewidth}
  \centering
  \includegraphics[width=1\linewidth]{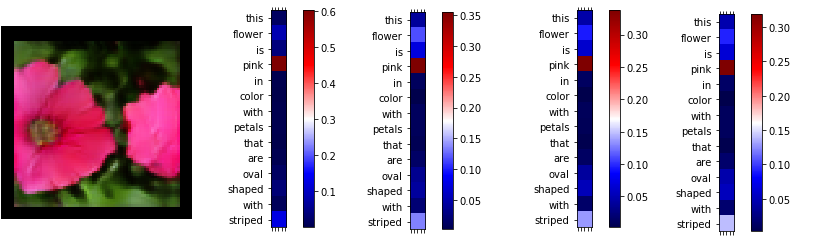}
  \caption{Text caption is: "The flower is pink i color with petals that are oval shaped wth striped."}
  \label{fig:sfig2}
\end{subfigure}
\centering
\caption{Attention weights for captions recorded for 4 timesteps. Colormap display how probable is the corresponding hidden state. 
Generated image is show on the left side.}
\label{fig:loss_curves}
\end{figure}